\documentclass[runningheads]{llncs}
\usepackage{graphicx}
\usepackage[table]{xcolor}

\usepackage{multirow}
\usepackage{booktabs}

\usepackage{hyperref}

\begin{document}
\title{Multi-scale Cell Instance Segmentation with Keypoint Graph based Bounding Boxes}
%
%
\author{Jingru Yi\inst{1} \and
Pengxiang Wu\inst{1} \and
Qiaoying Huang\inst{1} \and Hui Qu\inst{1} \and Bo Liu\inst{2} \and \\Daniel J. Hoeppner\inst{3} \and Dimitris N. Metaxas\inst{1}}
%
\authorrunning{J. Yi et al.}
%
\institute{Department of Computer Science, Rutgers University, Piscataway, NJ 08854, USA 
\email{jy486,pw241,qh55,hq43,dnm@cs.rutgers.edu}
\and
JD Digits, Mountain View, CA 94043, USA
\and
Lieber Institute for Brain Development, MD 21205, USA}
\maketitle              
\begin{abstract}
Most existing methods handle cell instance segmentation problems directly without relying on additional detection boxes. These methods generally fails to separate touching cells due to the lack of global understanding of the objects. In contrast, box-based instance segmentation solves this problem by combining object detection with segmentation. However, existing methods typically utilize anchor box-based detectors, which would lead to inferior instance segmentation performance due to the class imbalance issue. In this paper, we propose a new box-based cell instance segmentation method. In particular, we first detect the five pre-defined points of a cell via keypoints detection. Then we group these points according to a keypoint graph and subsequently extract the bounding box for each cell. Finally, cell segmentation is performed on feature maps within the bounding boxes. We validate our method on two cell datasets with distinct object shapes, and empirically demonstrate the superiority of our method compared to other instance segmentation techniques. Code is available at: \url{https://github.com/yijingru/KG\_Instance\_Segmentation}.

\end{abstract}
\section{Introduction}
Instance segmentation plays an important role in biomedical tasks such as cell migration study \cite{payer2018instance} and cell nuclei detection \cite{schmidt2018cell}. This problem requires not only classifying the objects, but also separating them from the neighboring instances. The main challenges in cell instance segmentation involve low contrast of cell boundaries, background impurities, cell adhesion and cell clustering.

To handle cell instance segmentation, one representative class of methods focus on segmenting the cell instances directly without the aid of bounding boxes. These box-free methods generally fail to separate the touching cells due to the lack of global understanding of the objects. For example, DCAN \cite{chen2016dcan} proposes to extract cell instances by overlapping their contours onto the semantic segmentation results. While being efficient, DCAN tends to produce over-segmentation due to the fuzzy contours between the touching cells.
STARDIST \cite{schmidt2018cell} suggests using convex polygons to separate cells, but with an assumption that the cell shape should be convex. CosineEmbedding \cite{payer2018instance} proposes to retrieve the cell instance by clustering the pixel embeddings. However, it tends to incur large number of false positives due to the separate clustering results for each individual cell.

To overcome the weakness of box-free instance segmentation, recent studies have sought to incorporate object detection into segmentation. These box-based methods first localize the cells via bounding boxes, and then perform individual cell segmentation on the regions defined by the bounding boxes. One major advantage of such methods is that they are able to distinguish cells based on their global object features instead of the local pixel-level information (e.g., boundary). As a result, box-based instance segmentation is more powerful in separating touching cells compared to the box-free strategies.

For box-based methods, a good object detector plays a critical role in the instance segmentation performance. However, previous methods (e.g., FCIS \cite{li2017fully} and Mask R-CNN \cite{he2017mask}) generally adopt anchor box-based detectors, which suffer from a severe imbalance between the number of positive and negative anchor boxes \cite{law2018cornernet}. Recently,  keypoints-based detectors are developed to solve the aforementioned problem. As one representative example, CornerNet \cite{law2018cornernet} proposes to detect the top-left and bottom-right points of an object for the generation of bounding box proposals, and achieves better accuracy than the anchor box-based detectors. However, such design also makes CornerNet prone to losing box proposals due to the missing detection of any corner points.


In this paper, we propose a new box-based cell instance segmentation method. In particular, we detect the top-left, top-right, bottom-left, bottom-right, and the center points of a cell rectangle using keypoints detection. Our motivation is that a bounding box can be represented by any three points or any pair of diagonal points among the five points. In this way, we effectively increase the probability of retrieving bounding boxes even when some keypoints are undetected. To generate bounding boxes, we group these points for each cell instance according to a keypoint graph. To further improve the detection accuracy, we use multi-scale feature maps to detect cells of different sizes. Cell segmentation is subsequently performed on feature maps cropped by the bounding box. We evaluate our method on two different cell datasets, and demonstrate its superiority in the instance segmentation of cells with different shapes.

\section{Method}
\label{method}
\begin{figure}[t!]
\includegraphics[width=\textwidth]{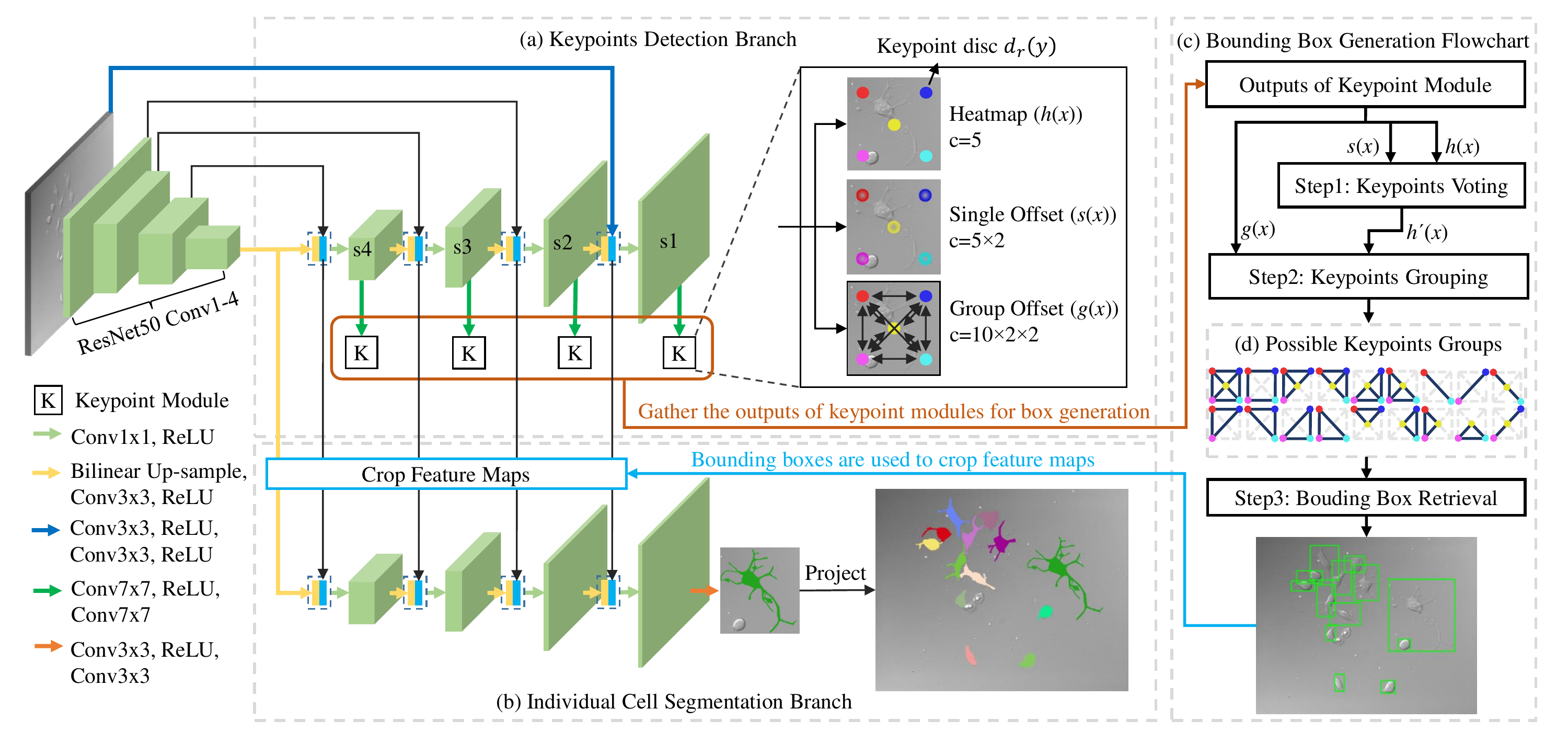}
\caption{Multi-scale cell instance segmentation framework. We use a ResNet-50 Conv1-4 \cite{he2016deep} as the backbone network. The framework contains two branches: (a) keypoints detection branch and (b) individual cell segmentation branch. The keypoint module outputs the heatmap $h(x)$, single offset map $s(x)$, and group offset map $g(x)$ that will be used for bounding box generation. $x$ represents a 2-D position in the map, $y$ is a 2-D position of the keypoint, $c$ indicates the channel of the map and $s$ denotes the scales. The red, blue, pink, green, yellow circles on these maps indicate the top-left, top-right, bottom-left, bottom-right and center points, respectively. (c) shows the bounding box generation flowchart, where $h^\prime (x)$ is the keypoint score map. (d) illustrates the possible keypoints groups that are used for box retrieval. } 
\label{fig1}
\end{figure}

The overview of our box-based instance segmentation framework is shown in Fig.~\ref{fig1}. We use a ResNet-50 Conv1-4 \cite{he2016deep} as the backbone network. The framework comprises two branches: keypoints detection branch (Fig.~\ref{fig1}a) and individual cell segmentation branch (Fig.~\ref{fig1}b). We illustrate the flowchart of generating cell bounding boxes in Fig.~\ref{fig1}c. 

\subsection{Bounding Box Generation}
\label{BBox_retrieval}
To obtain the bounding boxes of cells, we propose to detect the top-left, top-right, bottom-left, bottom-right, and the center points of a cell rectangle using keypoints detection. The keypoints detection branch is shown in Fig.~\ref{fig1}a, which outputs the heatmap $h(x)$, the single offset map $s(x)$ and the group offset map $g(x)$ at each scale $s_i,i=1,2,3,4$, for keypoints voting and grouping. $x$ represents a 2-D position (horizontal and vertical coordinates) in the image maps. Bounding boxes are then extracted according to the flowchart of Fig.~\ref{fig1}c.

\paragraph{Step1: Keypoints voting.} 
The keypoints voting takes two maps as input: the heatmap $h(x)$ and the single offset map $s(x)$. Heatmap is commonly applied in human pose estimation \cite{newell2016stacked,papandreou2018personlab} to predict the possibility of keypoints locations, which is a binary classification problem. To create the heatmap, we place a disc $d_r(y)=\{x:||x-y||\leq r\}$ around each keypoint $y$ (see Fig.~\ref{fig1}a), where $y$ denotes a 2-D position of a keypoint, and $r=5$ is the radius of the disc. The heatmap $h(x)$ contains $5$ channels (one channel per keypoint), where $h(x)=1$ for $x\in d_r(y)$, otherwise $h(x)=0$. We use binary cross entropy loss to optimize the parameters. After obtaining the heatmap of the keypoints, we use a single offset map $s(x)$ \cite{papandreou2018personlab} to extract the local maxima for each heatmap disc on $h(x)$. This can be viewed as a non-maximum suppression (NMS) operation. The single offset map $s(x)$ encodes the displacement between a keypoint $y$ and the points $x$ inside its disc:
\begin{equation}
  s(x)=y-x,~x\in d_r(y) . 
\end{equation}
The single offset map $s(x)$ contains $5\times 2$ channels (two channels per keypoint for displacements in the horizontal and vertical directions). We use $L_1$ loss to penalize the offset error. The gradients are only back-propagated inside the discs. The heatmap $h(x)$ and single offset map $s(x)$ are combined to generate the keypoint score map $h^{\prime}(x)$ via Hough voting using Hough accumulators \cite{papandreou2018personlab}:
\begin{equation}
    h^\prime(x) = \frac{1}{\pi r^2}\sum_{i=1}^N h(x_i)B(x_i+s(x_i)-x),
\end{equation}
where $x_i$ indexes the $i$-th 2-D position of the image, $B$ denotes the bilinear interpolation kernel. 

\paragraph{Step2: Keypoints grouping.}
The local maxima in the keypoint score map $h^\prime(x)$ represent the candidate positions of the keypoints. We apply a maximum filter to $h^\prime(x)$ and extract the keypoint locations via a peak threshold (0.004). After obtaining the keypoints, our next step is to group the keypoints for each cell instance. We propose a keypoint graph to group the keypoints, where the five types of keypoints are the vertices of the keypoint graph. We use a group offset to connect each pair of keypoints bi-directionally. In particular, for a pair of keypoints $(k,l)$ of a particular cell instance, the group offset from the $k$-th keypoint to the $l$-th keypoint is given by
\begin{equation}
    g_{k,l}(x)=(y_l-x), x\in d_r(y_k).
\end{equation}
The group offset map $g(x)$ has $10\times2\times2$ channels (two channels per pair of keypoints for displacements in the horizontal and vertical directions). The same to single offset map, we compute the $L_1$ loss to optimize the parameters and only back-propagate the loss at locations inside the keypoint discs. 
To group the keypoints, we first put all the detected keypoints into a queue and sort them according to their scores on $h^\prime(x)$. Then we pop the keypoint out of the queue in a descending order iteratively, and greedily connect the $(k,l)$ pair of keypoints using $g(x)$. At each iteration, we reject a repetitive detection by checking if the positions of two keypoints are within a disc.

\paragraph{Step3: Bounding box retrieval.} After aggregating the keypoint groups at scales $s_1,s_2,s_3,s_4$, our next step is to generate the bounding box for each cell instance. Fig.~\ref{fig1}d shows the possible keypoint groups that can be transformed to a full box. It can be seen that any three points or any pair of diagonal points in the keypoint graph can retrieve a box, which decreases the possibility of losing box proposals due to undetected points. We avoid detecting the same object mutiple times by applying NMS.

\subsection{Cell Segmentation}
\label{instance_segmentation}
After obtaining the bounding boxes for all cell instances in the input images, we perform the individual cell segmentation for each cell instance. Motivated by U-net \cite{ronneberger2015u}, we combine the feature maps from the shallow layers with the feature maps from the deep layers to take advantage of both high-level semantics and low-level image details. Specifically, we crop the multi-scale feature maps from the backbone network (see Fig.~\ref{fig1}b) and then perform a bottom-up segmentation for the cropped cell patchs. Note that we intentionally employ an individual cell segmentation branch (Fig.~\ref{fig1}b) for cell segmentation instead of directly reusing the feature map at $s_1$ (Fig.~\ref{fig1}a). Our motivation is to use the branch to guide the model to eliminate the interference from neighboring cells and learn an objectness concept especially for cells with irregular shapes (see Fig.~\ref{fig4}).

\section{Experiments}
\paragraph{Datasets.}
We evaluate our method on a neural cell dataset with irregular shapes and sizes and another cell nuclei dataset with regular shapes. 
The neural cell dataset contains 644 images that are sampled from a collection of time-lapse microscopic videos of rat CNS stem cells. The image size is $640\times 512$. We randomly select 386 image for training, 129 for validation and 129 for testing. For the cell nuclei dataset, we use the public training data of 2018 Data Science Bowl. This dataset is acquired under a variety of conditions and varies in the image size, cell type, magnification and imaging modality. From the total of 670 images, we randomly select 402 images for training, 134 images for validation and 134 images for testing. The input images are resized to $512\times 512$ in our experiments.

\paragraph{Implementation Details.}
We use the ground-truth bounding boxes to train the segmentation branch of Fig.~\ref{fig1}b. In testing, we perform the individual segmentation with the bounding boxes generated from keypoints detection. The training images are augmented using random expanding, cropping, flipping, contrast distortion and brightness distortion. We train the network for 100 epochs and stop when the validation loss does not decrease significantly. The weights of the backbone network are pre-trained from ImageNet. Other weights of the network are initialized from a standard Gaussian distribution. The model is implemented with PyTorch on NVIDIA K80 GPUs.

\paragraph{Evaluation Metrics.}
We use the average precision (AP)  at box-level IOU (intersection over union)  \cite{everingham2010pascal} at threshold of 0.5 and 0.7 to evaluate the detection performances. We use the AP at mask-level IOU \cite{he2017mask,li2017fully} at threshold of 0.5 and 0.7 to evaluate the instance segmentation performances. We also report the mean mask-level IOU \cite{yi2019attentive} between the predicted segmentation masks and the ground truth masks at threshold of 0.5 and 0.7.

\begin{table}[t!]
\centering
\scriptsize
\caption{Cell instance segmentation evaluation results. Seg $s_1$ means directly performing individual cell segmentation from feature map $s_1$. Seg branch refers to the individual cell segmentation branch in Fig.~\ref{fig1}b. }
\label{tab2}
\setlength\extrarowheight{1pt}
\begin{tabular}{l|cccc|cccc}
\hline
\multirow{2}{*}{Model} & \multicolumn{4}{c|}{Neural Cell} & \multicolumn{4}{c}{DSB2018}\\
  & AP@0.5 & AP@0.7 & IOU@0.5 & IOU@0.7 & AP@0.5 & AP@0.7 & IOU@0.5 & IOU@0.7\\
\hline
DCAN \cite{chen2016dcan} & 45.03 &10.76 & 64.49 & 75.91 &51.88   & 23.45 & 74.08&  82.56 \\
CosineEmbedding \cite{payer2018instance}&25.93 & 9.09 & 62.22 & 75.07 & 17.87 & 3.41&64.14&76.84 \\
Mask R-CNN \cite{he2017mask}& 66.02 & 32.10 & 72.10 & 79.30 &  69.88& 54.69 & 80.57 & 84.83\\
Ours (seg $s_1$)&  78.49 & 50.97& 75.51&\textbf{80.42 }& 71.38 & 59.38& 83.10& 86.10\\
Ours (seg branch)&  \textbf{88.03} & \textbf{ 63.08}& \textbf{77.04}& 79.94& \textbf{71.58 } & \textbf{59.81}& \textbf{83.29}& \textbf{86.22}\\
\hline
\end{tabular}
\end{table}

\begin{figure}[t!]
\includegraphics[width=\textwidth]{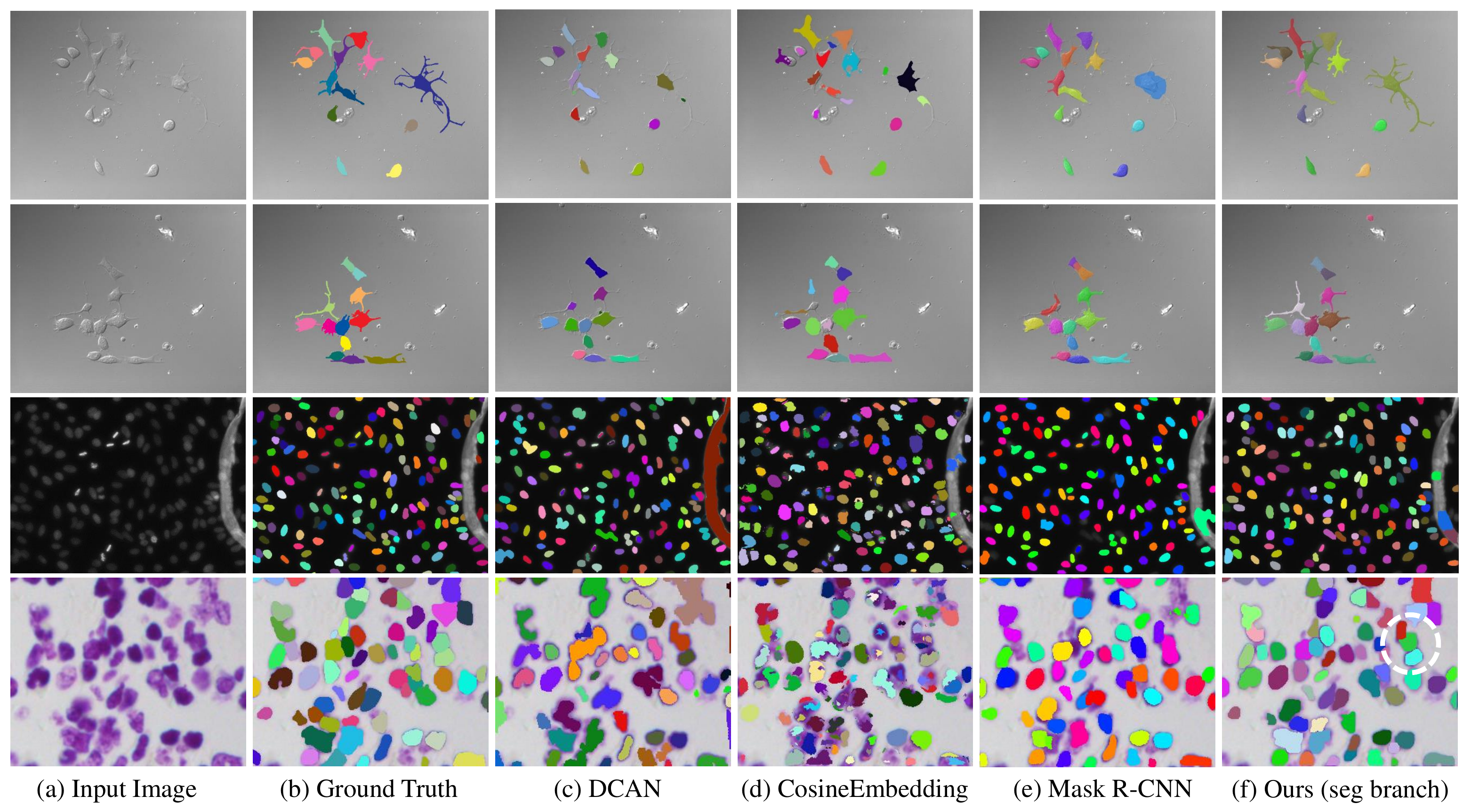}
\caption{Qualitative cell instance segmentation results on neural cells (top two rows) and cell nuclei (bottom two rows). We compare our instance segmentation method with DCAN \cite{chen2016dcan}, CosineEmbedding \cite{payer2018instance} and Mask R-CNN \cite{he2017mask}. The white dotted circle shows an example where our method separates the touching cells.} 
\label{fig3}
\end{figure}

\begin{figure}[t!]
\includegraphics[width=\textwidth]{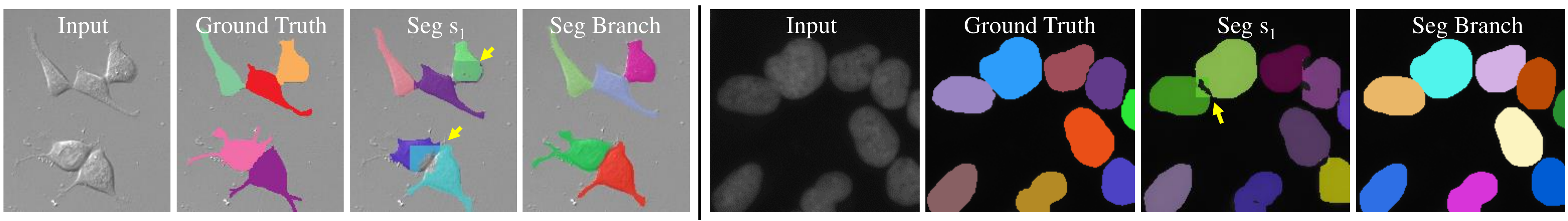}
\caption{Comparison between individual cell segmentation from feature map $s_1$ (seg $s_1$) and from individual cell segmentation branch (seg branch).  The left four columns are neural cells. The right four columns are cell nuclei. The yellow arrows point to the over-segmentions of method seg $s_1$.} 
\label{fig4}
\end{figure}

\section{Results and Discussion}
We compare our instance segmentation method with DCAN \cite{chen2016dcan}, CosineEmbedding \cite{payer2018instance} and Mask R-CNN \cite{he2017mask}. The quantitative and qualitative results are reported in Table~\ref{tab2} and Fig.~\ref{fig3}. As can be seen from Fig.~\ref{fig3}, DCAN tends to remove the details along with the cell boundaries for neural cells. For nuclei dataset, it is unable to differentiate the touching cells due to the unclear cell boundaries. CosineEmbedding \cite{payer2018instance} clusters the pixel embeddings to segment the cell instances. However, the clustering usually generates multiple separate clusters for the same cell instance. Therefore, it suffers from huge false positives and achieves inferior performance in detection, especially for crowded nuclei dataset (Table \ref{tab1}). Mask R-CNN \cite{he2017mask} is superior in cell detection, but it cannot predict the long and slender structures of the cells because of its ROI align mechanism. Compared to these methods, our keypoints detection-based cell instance segmentation performs well in both capturing the long and slender cell structures and separating the touching cells. Moreover, from the last two rows of Table \ref{tab2} and from Fig.~\ref{fig4} we can observe that the individual cell segmentation branch (Fig.~\ref{fig1}b) performs better than segmentation based solely on feature map $s_1$ (Fig.~\ref{fig1}a), especially for neural cells. The reason would be that the the model lacks an object concept for cells when segmenting them only using feature map $s_1$. As a result, it is difficult for the model to filter out the interference of neighboring cells (see Fig.~\ref{fig4}). In contrast, the individual cell segmentation branch is able to provide guidance for the network to eliminate the unrelated cell parts for each cell ROI patch.

\begin{table}[t!]
\centering
\caption{Detection evaluation results. Single-scale dec means the keypoints detection at $s_1$ (see Fig.~\ref{fig1}), while multi-scale dec means the detection at $s_1, s_2, s_3, s_4$. }\label{tab1}
\setlength\extrarowheight{1pt}
\setlength{\tabcolsep}{8pt}
\scriptsize
\begin{tabular}{l |c c| c c}
\hline
\multirow{2}{*}{Model} & \multicolumn{2}{c|}{Neural Cell} & \multicolumn{2}{c}{DSB2018}\\ 
 & AP@0.5 & AP@0.7 & AP@0.5 & AP@0.7\\
\hline
DCAN \cite{chen2016dcan} & 13.85& 9.09 &  52.86 & 31.02  \\  
CosineEmbedding \cite{payer2018instance}&  27.45 & 10.99 & 11.93&1.30 \\
Mask R-CNN \cite{he2017mask} &64.65&17.76 &  69.93& 45.25 \\
CornerNet \cite{law2018cornernet} & 60.42 & 39.75 & 47.99& 38.35\\ 
Ours (single-scale dec)& 60.97& 46.69&  \textbf{80.39}& \textbf{69.11}\\
Ours (multi-scale dec)&\textbf{79.30} & \textbf{55.18} &  80.14& 67.60\\
\hline
\end{tabular}
\end{table}

\begin{figure}[t!]
\includegraphics[width=\textwidth]{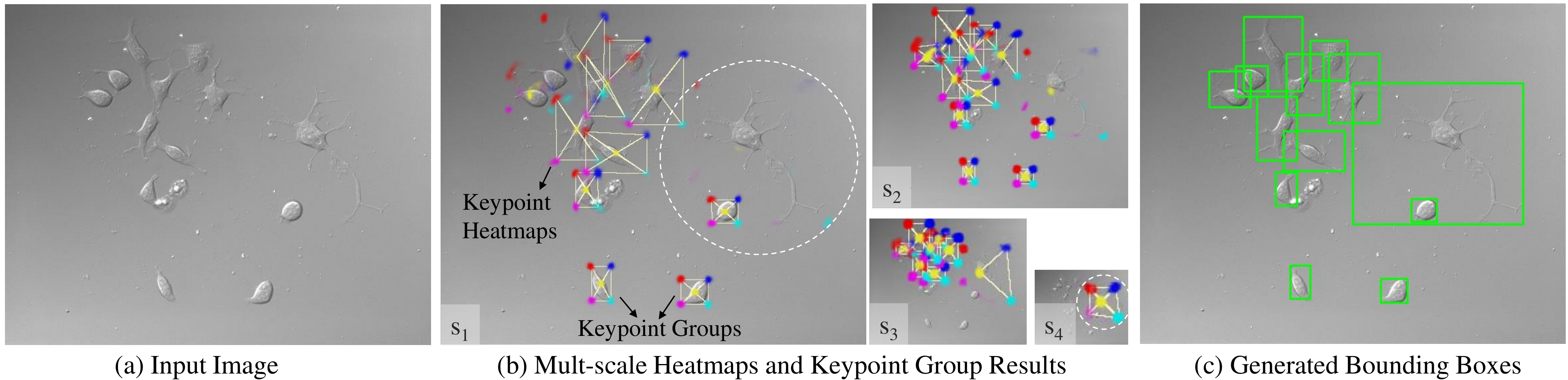}
\caption{Visualization of heatmap predictions and keypoint groups overlaid on the input images. We show the heatmaps at four scales $s_i, i=1,2,3,4$. The circles illustrate an example that a large cell is unrecognized at scale $s_1$ but is captured at scale $s_4$.} 
\label{fig2}
\end{figure}

We also report the cell detection comparison results in Table \ref{tab1} to analyze the detection ability of our keypoint graph-based detection. We add the comparison between our method and the keypoint-based detector CornerNet \cite{law2018cornernet} in Table \ref{tab1}. It can be seen that our keypoint graph-based detector achieves better accuracy in capturing the bounding boxes of the cells, compared to the other methods.  Besides, the multi-scale detection performs better than single-scale detection for neural cell dataset. To illustrate the reason, we visualize the heatmap predictions and the keypoint groups in Fig.~\ref{fig2}. As can be seen, the model can hardly detect the keypoints of cells with large sizes on the shallow layers. One possible reason would be that the shallow layer has a  small receptive field, and thus it is difficult for the model to recognize a large object on the shallow layers. This defect also brings difficulty in predicting the correct displacement between two keypoints pairs for large cells, due to the loss of objectness concept. Compared to the shallow layers, the deep layers are able to detect the large cells because of their large receptive fields, as shown in Fig.~\ref{fig2}. For cell nuclei, we do not observe obvious superiority for multi-scale cell detections since the sizes of nuclei are at a similar scale.

\section{Conclusion}
In this paper, we propose a new instance segmentation method that combines the keypoint-based detector with the individual cell segmentation. In particular, we propose a novel keypoint-based detector that is more effective in generating bounding box proposals. The experimental results demonstrate the advantages of our method in segmenting the cell instances with both regular and irregular shapes, compared to the other instance segmentation methods.

%
%
%
%

\end{document}